\documentclass[conference]{IEEEtran}
\usepackage{cite}
\usepackage{amsmath,amssymb,amsfonts}
\usepackage{algorithm}
\usepackage{algorithmic}
\usepackage{graphicx}
\usepackage{textcomp}
\usepackage{xcolor,comment}
\usepackage{color}
\usepackage{multirow,multicol}

\usepackage[compact]{titlesec}
\titlespacing{\section}{0pt}{0ex}{0ex}
\titlespacing{\subsection}{0pt}{0ex}{0ex}
\titlespacing{\subsubsection}{0pt}{0ex}{0ex}

\ifCLASSINFOpdf
\else
\fi
\newcommand{\authone}[1]{\textcolor{black}{#1}}  
\newcommand{\authtwo}[1]{\textcolor{black}{#1}}  
\begin{document}
%
\title{\mbox{SDN}-Based False Data Detection With Its Mitigation and Machine Learning Robustness for In-Vehicle Networks}

\author{
\IEEEauthorblockN{Kaiqi Xiong and Yi Li}
\IEEEauthorblockA{University of South Florida\\
Tampa, FL 33620 USA\\
Email: xiongk@usf.edu and lyirene13@hotmail.com}
}

\author{
\IEEEauthorblockN{Long Dang, Thushari Hapuarachchi, Kaiqi Xiong, and Yi Li}
\IEEEauthorblockA{
ICNS Lab and Cyber Florida, University of South Florida, Tampa, FL 33620, USA \\
Emails: \{longdang, saumya2, xiongk\}@usf.edu and lyirene13@hotmail.com
}
}

%


\maketitle

\begin{abstract}
As the development of autonomous and connected vehicles advances, the complexity of modern vehicles increases, with numerous Electronic Control Units (ECUs) integrated into the system. In an in-vehicle network, these ECUs communicate with one another using a standard protocol called Controller Area Network (CAN). Securing communication among ECUs plays a vital role in maintaining the safety and security of the vehicle. This paper proposes a robust SDN-based False Data Detection and Mitigation System (FDDMS) for in-vehicle networks. Leveraging the unique capabilities of Software-Defined Networking (SDN), FDDMS is designed to monitor and detect false data injection attacks in real-time. Specifically, we focus on brake-related ECUs within an SDN-enabled in-vehicle network. First, we decode raw CAN data to create an attack model that illustrates how false data can be injected into the system. Then, FDDMS, incorporating a Long Short-Term Memory (LSTM)-based detection model, is used to identify false data injection attacks. We further propose an effective variant of the DeepFool attack to evaluate the model's robustness. To countermeasure the impacts of four adversarial attacks including Fast gradient descent method, Basic iterative method, DeepFool, and the DeepFool variant, we further enhance a re-training technique method with a threshold based selection strategy. Finally, a mitigation scheme is implemented to redirect attack traffic by dynamically updating flow rules through SDN. Our experimental results show that the proposed FDDMS is robust against adversarial attacks and effectively detects and mitigates false data injection attacks in real-time.
\end{abstract}


%
\IEEEpeerreviewmaketitle

\section{Introduction}\label{sec:introduction}In modern vehicles, especially connected and autonomous vehicles (CAVs), the number of Electronic Control Units (ECUs) has significantly increased. The Controller Area Network (CAN), which is a standard communication protocol used in the vehicle network, enables ECUs to communicate with each other~\cite{li2020countermeasures}. CAVs leverage additional ECUs to support advanced driving tasks such as lane change, parking assistance, automatic emergency braking, and more~\cite{mudalige2015efficient}. To enable these features, the vehicle needs to communicate not only within itself but also with other vehicles and infrastructure, such as in Vehicle-to-Vehicle (V2V) and Vehicle-to-Infrastructure (V2I) communication~\cite{wang2019survey}.

Protecting the in-vehicle network is just as critical as protecting the security of Vehicle-to-Everything (V2X) communication. Modern vehicles can have over 70 ECUs, and research has shown that these ECUs are vulnerable to remote attacks~\cite{huang2018vehicle, avatefipour2018state}. \authone{Because ECUs are interconnected, an attacker can gain remote access to the in-vehicle network via wireless networks such as WLAN or Bluetooth, and compromise ECUs}. As a result, vehicle networks are susceptible to a variety of attacks, including intrusion attacks and False Data Injection Attacks (FDIA)~\cite{carsten2015vehicle}.
Software-defined Networking (SDN), well known for its flexibility and programmability, can separate a network traffic controller from the data plane in a network~\cite{kreutz2015software}. Using the SDN based controller provides better traffic flow monitoring and forwarding for network management. In the context of a vehicle network, detecting FDIA in real time is crucial to prevent potentially catastrophic outcomes like traffic accidents~\cite{islam2018cybersecurity}. The use of SDN allows for more effective detection and mitigation of attack traffic in real time.


FDIA can cause significant damage, as seen in the 2015 Ukraine blackout~\cite{liang20162015}. In vehicle networks, such attacks could lead to life-threatening situations, like car crashes, if not detected. To address this, we propose an SDN-based False Data Detection and Mitigation System (FDDMS) for in-vehicle networks. The system uses SDN to detect and mitigate FDIA on ECUs in real time. It consists of two main components: a False Data Detection Module with Long Short-Term Memory (LSTM) for detection, and a False Data Mitigation Module. The SDN controller monitors network traffic and, upon detecting an attack, updates forwarding rules in OpenFlow switches to block the attack.

\authone{While the LSTM based module can detect FDIA signals with high accuracy, recent studies have shown that neural network based models are vulnerable to adversarial attacks~\cite{DBLP:journals/corr/abs-2112-02797, GoodfellowFGSM, DBLP:journals/corr/KurakinGB16, Moosavi-Dezfooli16}. Thus, FDIA signals, when combined with adversarial attacks, can cause the LSTM model to misclassify them as "Normal" instead of "Attack". FGSM generates adversarial examples by adding gradient based perturbations~\cite{GoodfellowFGSM}. The FGSM based perturbations have the same direction as the gradients of the loss function. Because FGSM is designed to be fast, its perturbations are often large. BIM~\cite{DBLP:journals/corr/KurakinGB16}, an iterative version of FGSM, finds more subtle perturbations compared to FGSM. Using BIM to craft adversarial FDIA signals requires hyperparameter tuning to minimize distortion and maintain a high attack success rate (ASR). Thus, both approaches suffer from a trade-off between minimal perturbations and ASR~\cite{cerracchio2024investigating}. Meanwhile, DeepFool was specifically designed to find the minimal perturbation for misclassifications~\cite{Moosavi-Dezfooli16}. Because DeepFool relies on iteratively approximating the decision boundary of a non-linear model, it can be computationally intensive, especially for complex models such as the LSTM model with recurrent units~\cite{li2020countermeasures} and datasets with multiple classes. Thus, while these attacks significantly reduce the detection accuracy of the model with relatively small perturbations to the inputs, it is still crucial to study of the robustness of the LSTM model and the potential impact of adversarial attacks especially for in-vehicle networks. In this study, we advance the study of machine learning robustness by improving the DeepFool attack. Compared to the original version, we propose a practical variant that can significantly reduce the distortion of FDIA signals while maintaining high ASR, with a potential decrease in computational costs.}

\authone{To protect neural networks from the adversarial attacks, Madry et al.~\cite{MadryMSTV17} proposed an effective training technique called adversarial training (AdvTrain). Ilyas et al.~\cite{ilyas2019adversarial} showed that AdvTrain could improve several image classfiers's robustness against adversarial attacks by removing non-robust features from adversarial inputs and forcing the classifiers to learn only robust features. While AdvTrain has been shown to be effective against advanced adversarial attacks~\cite{athalye2018obfuscated, lin2020secure, xie2020smooth}, it suffers from the trade-off between natural accuracy and robust accuracy~\cite{zhang2019theoretically, xie2020smooth}. A "normal" accuracy is the test accuracy of the trained model performed on the clean data. A "robust" accuracy is the test accuracy of the trained model performed on the adversarial data. Thus, we adopted an effective AdvTrain technique proposed in~\cite{li2021adversarial} that can train a robust detection model against four adversarial attacks including FGSM, BIM, DeepFool, and our DeepFool variant attacks. Compared to~\cite{MadryMSTV17}, Li.et al's iterative re-training technique~\cite{li2021adversarial} involves 1) gradually expanding the training set size with adversarial examples and 2) re-training the detection model using both adversarial examples and original training examples sampled from the expanded training dataset. We further extend~\cite{li2021adversarial} by introducing a score and a threshold parameter to select challenging adversarial examples for the model to learn. The score is based on the confidence of the model's prediction for each adversarial examples and its ground truth labels. We skip adversarial examples with a higher confidence score above the predefined threshold because the higher scores might indicate that the model is already certain about its predictions. To optimize computational costs, we also apply an early stopping technique to check the robust accuracy against the FGSM attack on a validation set.}

\authone{We conducted our experiments on a dataset of real in-vehicle CAN bus~\cite{lee2017otids} to evaluate the efficacy of the proposed FDDMS. In particular, we assumed a white-box scenario that was used in~\cite{li2021adversarial,cerracchio2024investigating}. This assumption allows calculating the gradients needed to create effective adversarial examples with minimal distortion to the FDIA signal. Because we consider detecting adversarial inputs as a binary classification problem, we focused on the untargeted versions of FGSM, BIM, and DeepFool. In a binary classification set-up, our DeepFool attack only needs to find the closest approximated decision boundary that divide the two classes, which might require a smaller perturbation and less computational costs than a multi-classification problem. Our DeepFool variant achieved the lowest $L_{0}, L_{2},$ and $L_{\infty}$-norm and a comparable high ASR of 99\% when compared to FGSM, BIM, and the original DeepFool. We found that the LSTM based model trained on challenging FGSM examples achieved the best normal accuracy of 99.47\%, and the best robust accuracy. The robust accuracy against FGSM, BIM, the original DeepFool, and our DeepFool attacks was 99.475\%, 99.47\%, 98.95\%, and 99.47\%, respectively. Finally, the SDN and mitigation scheme took a low average time of 5.3617 ms, 1.6217 ms, and 0.7991 ms for message transmission, adversarial detection, and adversarial mitigation.}
In summary, our study makes the following key contributions.
\begin{itemize}
\item We design and implement an efficient variant of the DeepFool attack that can effectively reduce the distortion of adversarial perturbations added to False Data Injection Attack samples while maintaining a high attack success rate.
\item We design an efficient re-training technique that can select difficult adversarial samples for model re-training by using a score based mechanism. Furthermore, we showed that our adversarially trained models had a minimal trade-off between normal and robust accuracy.
\item To redirect attack traffic, we design and implement a mitigation scheme by dynamically updating flow rules through SDN.
\end{itemize}

The rest of the paper is organized as follows. Section~\ref{sec:related} reviews related work on FDIA and in-vehicle network security. Section~\ref{sec:methodology} presents the design of the proposed FDDMS. Section~\ref{sec:evaluation} evaluates the system's performance, including detection accuracy. Finally, Section~\ref{sec:conclusion} concludes the paper and discusses future directions for research.
\section{Related Work}\label{sec:related}The characteristics of CAN data frames including its broadcast communication, lack of authentication, and absence of encryption make an in-vehicle network vulnerable to a variety of attacks. Attackers can exploit interfaces such as Bluetooth and Wi-Fi to infiltrate the in-vehicle network and execute attacks such as Denial of Service (DoS), replay attacks, and DIA~\cite{liu2017vehicle,pham2021survey}.

M{\"u}ter et al.\cite{muter2010structured} proposed an anomaly detection system for in-vehicle networks that leverages the unique characteristics of the CAN bus. Their system used eight attack detection sensors to identify anomalies in the network. Later, M{\"u}ter et al.\cite{muter2011entropy} enhanced this approach by introducing an entropy-based anomaly detection technique, which added a reactive layer of protection for the CAN bus. Their method was shown to be effective in detecting a range of attack scenarios in real-world vehicle settings. Other studies and surveys include~\cite{zhao2024potential,baccari2024anomaly,fang2024anomaly}.

Cho et al.~\cite{cho2016fingerprinting} developed a clock-based intrusion detection system (CIDS) to safeguard in-vehicle networks. CIDS works by fingerprinting Electronic Control Units (ECUs) based on the intervals between periodic CAN messages. A baseline of normal clock behavior was created using the recursive least squares algorithm, which is then used to detect intrusions. They considered three types of attack scenarios: fabrication, suspension, and masquerade attacks, and found that CIDS was effective at identifying intrusions in the in-vehicle network.

Recently, there has been a growing use of machine learning based intrusion detections techniques for in-vehicle networks. Kang et al.\cite{kang2016intrusion} proposed a deep learning based intrusion detection system (IDS). The model achieved better accuracy in detecting intrusions by using a pre-trained unsupervised deep belief network for weight initialization. Kuwahara et al.\cite{kuwahara2018supervised} utilized both supervised and unsupervised learning approaches for intrusion detection with sequential data formed by processing CAN messages within fixed time windows.

FDIA, first introduced by Liu et al.\cite{liu2011false} to disrupt state estimation in power systems, have since been explored in domains like smart grids\cite{guan2015comprehensive}, wireless sensor networks~\cite{jeba2012false}, and vehicle networks~\cite{cao2008proof}.
Cao et al.\cite{cao2008proof} proposed the Proof-of-Relevance framework for defending against FDIA in VANETs, relying on consensus from witness vehicles to prevent false data propagation. Moore et al.\cite{moore2017modeling} identified signal frequency as a key feature for CAN bus intrusion detection, detecting FDIA by observing irregular refresh rates in regularly transmitted PID signals. Their method showed promising results across multiple attack scenarios.

The concept of Software-Defined Vehicular Networks (SDVN) has gained significant attention due to the potential advantages of SDN in network management~\cite{he2016sdvn}. Singh et al.~\cite{singh2018ml} proposed a machine learning-based system for detecting Distributed Denial of Service (DDoS) attacks in Vehicle-to-Infrastructure (V2I) communications. Their system employed several machine learning models to identify the most effective approach for detecting DDoS attacks. Khan et al.~\cite{khan2019vehicle} presented an effective framework for detecting and mitigating in-vehicle false information attacks using a combination of machine learning and SDN.

\section{Methodology}\label{sec:methodology}CAN is the industry standard for in-vehicle network communication because it can facilitate efficient data exchange for complex vehicle operations. CAN utilizes a broadcasting method for transmitting messages. When an ECU wants to send data, it transmits a CAN message onto the bus. The message becomes available to all other ECUs connected to the same CAN bus. Unlike some other network protocols, CAN messages do not contain the source and destination information. Thus, it presents a significant security challenge because an attacker can easily inject false data into the CAN bus without being detected, leading to vehicle accidents~\cite{li2020countermeasures}. 

The decoding and data processing technique used in this study is first describe in this section. Next, we extend the attack model with adversarial attacks. Specifically, we present our DeepFool variant for generating adversarial perturbations that bypass the detection model with smaller perturbation. Lastly, we show our SDN-based False Data Detection and Mitigation System (FDDMS) for in-vehicle networks. We further improve the detection module by introducing an effective adversarial-retraining technique. The technique can protect the detection model against four adversarial attacks including FGSM, BIM, the original DeepFool, and the proposed DeepFool variant.\vspace{-1.8mm}
\subsection{Decoding and Data Processing}
In this paper, we used the dataset from real a KIA SOUL in-vechicle CAN bus~\cite{lee2017otids} to thoroughly investigate the efficacy of proposed FDDMS. The dataset comprises of an attack free state data and three other data injected with denial of service attacks, fuzzy attacks, and impersonation attacks. The attack free file has 3,713,146 rows and four columns including ID, DLC, Data, and Timestamp. The Timestamp column records approximately 1,904 seconds of data. We only used the attack free state data because we want to implement a false data injection using our attack model described in the following section. \vspace{-5.0mm}
\begin{figure}[t]
\small
	\centering
	\includegraphics[width=\columnwidth]{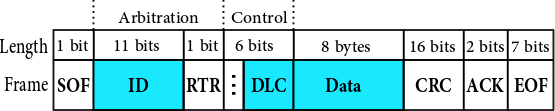}  
	\caption{Format of a CAN data frame}
	\label{fig:canframe}
\end{figure}

\begin{table}[h]
    \caption{Details For Decoding Raw CAN Data}
    \small\addtolength{\tabcolsep}{-1.6pt}
    \footnotesize
    \begin{tabular}{| l | l | l | l | l | l | l | l |} 
    \hline
      {\textbf{No.}} &{\textbf{Signal}}& {\textbf{MSG}}  & {\textbf{CID}} & \textbf{MID} & \textbf{Bits} & \textbf{Scale} & \textbf{Offset} \\ \hline
      1  & SAS\_Angle    & SAS11 & 02b0 & 688 & 0-15  & 0.10 & 0.00\\ \hline
      2  & SAS\_Speed    & SAS11 & 02b0 & 688 & 16-23 & 4.00 & 0.00\\ \hline
      3  & MsgCount      & SAS11 & 02b0 & 688 & 32-35 & 1.00 & 0.00\\ \hline
      4  & CheckSum      & SAS11 & 02b0 & 688 & 36-39 & 1.00 & 0.00\\ \hline
      5  & TQ\_COR\_STAT & EMS11 & 0316 & 790 & 4-5   & 1.00 & 0.00\\ \hline
      6  & TQI\_ACOR     & EMS11 & 0316 & 790 & 8-15  & 0.39 & 0.00\\ \hline
      7  & N             & EMS11 & 0316 & 790 & 16-31 & 0.25 & 0.00\\ \hline
      8  & TQI           & EMS11 & 0316 & 790 & 32-39 & 0.39 & 0.00\\ \hline
      9  & TQFR          & EMS11 & 0316 & 790 & 40-47 & 0.39 & 0.00\\ \hline
      10 & VS            & EMS11 & 0316 & 790 & 48-55 & 1.00 & 0.00\\ \hline
      11 & MUL\_CODE     & EMS12 & 0329 & 809 & 6-7   & 1.00 & 0.00\\ \hline
      12 & TEMP\_ENG     & EMS12 & 0329 & 809 & 8-15  & 0.75 & -48.00\\ \hline
      13 & BRAKE\_ACT    & EMS12 & 0329 & 809 & 32-33 & 1.00 & 0.00\\ \hline
      14 & TPS           & EMS12 & 0329 & 809 & 40-47 & 0.47 & -15.02\\ \hline
      15 & PV\_AV\_CAN   & EMS12 & 0329 & 809 & 48-55 & 0.39 & 0.00\\ \hline
      16 & VB            & EMS14 & 0545 & 1349& 24-31 & 0.10 & 0.00\\ \hline
      17 & TQI\_MIN      & EMS16 & 0260 & 608 & 0-7   & 0.39 & 0.00\\ \hline
      18 & TQI           & EMS16 & 0260 & 608 & 8-15  & 0.39 & 0.00\\ \hline
      19 & TQI\_TARGET   & EMS16 & 0260 & 608 & 16-23 & 0.39 & 0.00\\ \hline
      20 & TQI\_MAX      & EMS16 & 0260 & 608 & 40-47 & 0.39 & 0.00\\ \hline
    \end{tabular}
    \label{tab:signal}
\end{table}\vspace{-1.8mm}
For an in-vehicle network, the CAN data frames are used by ECUs to transmit messages and data to other ECUs on the bus~\cite{li2021adversarial}. Figure~\ref{fig:canframe} shows the format of a CAN data frame. The CAN data frame contains three critical fields including Arbitration, Control, and Data. The Arbitration field contains the CAN IDs (CID), which identifies the associated ECU~\cite{cerracchio2024investigating}. Next, the Control indicates the length of the data field~\cite{li2020countermeasures}. Lastly, the 8-bytes Data carries the signals being transmitted~\cite{cerracchio2024investigating}. To obtain decimal signal values, we decoded the raw CAN data frames using a generic KIA DBC file decoder~\cite{opendbc}. The DBC file contains the message IDs (MSG), the CIDs, the starting bit position of signals, the length in bits of signals, the scale and offset values to attain human-readable signal values from raw bits of data. Table~\ref{tab:signal} provides all the necessary details for decoding the raw CAN data frames.  

Each CID is linked to a specific ECU while each ECU has multiple signals~\cite{li2021adversarial}. We examined 20 signals related to five ECUs, as described in~\cite{khan2019vehicle,li2021adversarial}. These ECUs are responsible for controlling various critical vehicle functions, ranging from Anti-lock Braking System (ABS), Electronic Parking Brake System (EPB), Electronic Stability Control System (ESC) for braking, and Motor Driven Power Steering (MDPS) for steering to Engine Management System (EMS). MDPS and EMS ECUs send CAN messages/frames to ABS, EPB, ESC. EMS broadcasts four types of CAN messages (MSG) including EMS11, EMS12, EMS14, and EMS16 while MDPS ECU broadcasts SAS11~\cite{li2021adversarial}. We began the decoding process by converting the CIDs in hexadecimal representation to match the MIDs in decimals. Then, we identified signal bits within the 8-byte data field. 
Once we identified the bit range for a signal, we could extract these bits from the 8-byte data field. Lastly, we used Equation~\ref{Eq:V_signal} to convert the raw binary values into decimal values:\vspace{-1.8mm}
\begin{equation}
V_{signal} = o + s \times V_{raw},
\label{Eq:V_signal}
\end{equation}
where $V_{signal}$ denotes the signal value after decoding, $s$ is the scale value, $o$ is the offset  from the DBC file, and $V_{raw}$ is the raw decimal data value from the data field.
%
\begin{figure}[t]
	\centering
	\includegraphics[width=0.95\columnwidth]{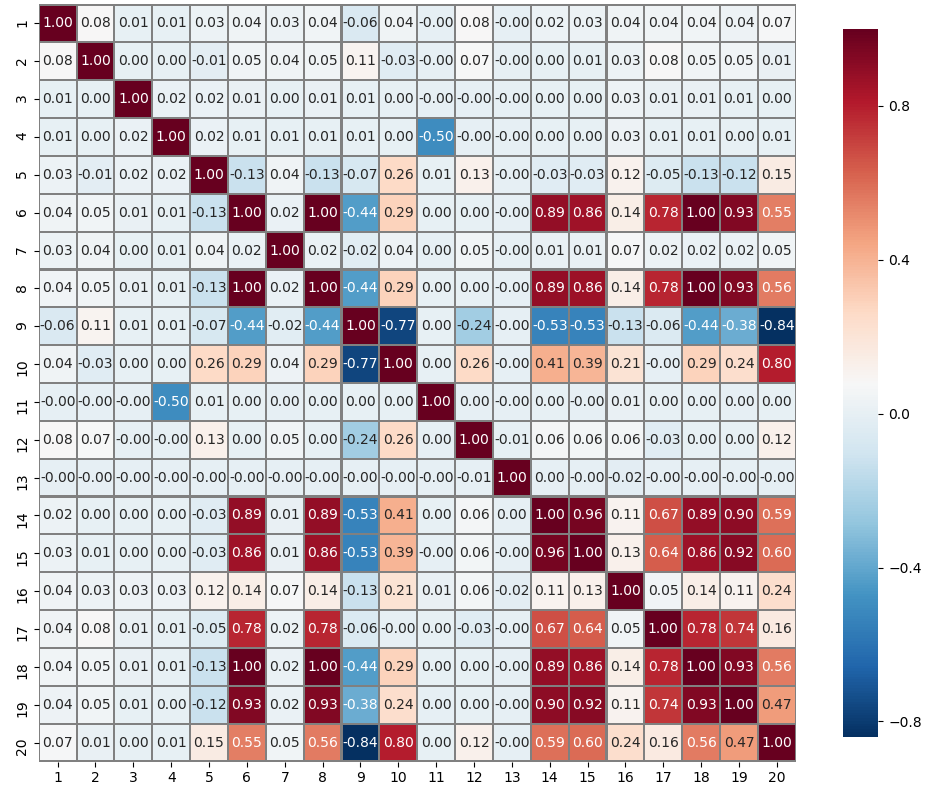}  
	\caption{Heatmap of correlation matrix among all features}
	\label{fig:heatmap}
\end{figure}

Given that our analysis focused on the five brake-related ECUs, we filtered all irrelevant data. We retained 1,904 seconds of data and 952,101 rows of CAN data frames, with a relatively uniform distribution of frames across different CIDs. We then selected the 20 signals outlined in Table~\ref{tab:signal} as features to build the training, validation, and test datasets.

\subsection{Attack Model}
In this section, we present our FDIA model for in-vehicle networks and demonstrate how false data is injected into the attack-free dataset.
\subsubsection{\textbf{Correlation Analysis}}
Before injecting false data, we identified the relationships between the features. Modifying the value of one feature can impact other features that are highly correlated with it. Therefore, we began by conducting a correlation analysis across all 20 features. We used Pearson's Correlation Coefficient to assess the linear correlation between pairs of features. Figure~\ref{fig:heatmap} displays the correlation heatmap for all features. Based on our qualitative analysis of the dataset, we set the significance threshold for correlation at 0.75 to determine whether two features were considered correlated.
For instance, TQI\_ACOR shows strong correlations with the following features: TQI (0.999537), TPS (0.887654), PV\_AV\_CAN (0.862707), TQI\_MIN (0.778768), and TQI\_TARGET (0.934837).
\begin{figure}[t]
	\centering
	\includegraphics[width=\columnwidth]{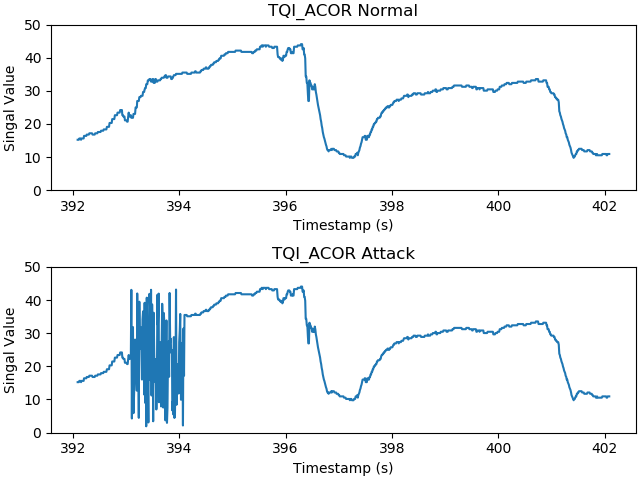}  
	\caption{Attack on TQI\_MAX signal}
	\label{fig:atkExample}
\end{figure}
\subsubsection{\textbf{Attack Model}}
The reduced dataset contained 1,904 seconds of data and 952,101 rows of CAN data frames. The dataset was divided based on timestamps, with each instance spanning 10 seconds. For example, the first instance covered seconds 1 to 10, the second instance spans seconds 2 to 11, and so on. In total, 1,894 instances were created. Of these instances, half were randomly selected for false data injection.

Based on the analysis of feature correlations, we selected the following signals/features to target in the attack: TQI\_ACOR, N, TQFR, and VB. Because these signals are also correlated with other features, we ultimately chose 11 signals for false data injection. These signals were derived from the EMS11, EMS12, EMS14, and EMS16 message frames, and are associated with the braking and Engine Management System.

An attacker can easily manipulate the original data by injecting attack data, which can be generated in various ways. 
In this paper, we first analyzed the normal range for each signal using the DBC file for the KIA Soul. We then generated random values following a uniform distribution within the defined normal range. For each 10-second instance, we randomly selected one second during which to inject the attack. The equation for our attack model is shown as follows:\vspace{-1.8mm}
\begin{equation}
V_{attack}^i = V_{normal}^i + \delta^i, \,
\end{equation}
where 
$\delta^i \sim \mathcal{U}(-V_{normal}^i,\,V_{max}^i - V_{normal}^i)$ represents a uniformly distributed random value added to the normal data to generate false data. Here, $i$ denotes the signal index, $V_{attack}$ is the false data injected by the attacker, and $V_{max}^i$ is the maximum value of the signal. For example, $V_{attack}^6$ refers to the false data injected into signal 6, which corresponds to TQI\_ACOR. $V_{normal}$ represents the normal data without any attack.
%
Figure~\ref{fig:atkExample} illustrates an example of false data injection into the TQI\_ACOR signal using the attack model, along with the normal signal values. The plot shows the TQI\_ACOR normal and attack signal values over the time interval from 392 to 402 seconds. The false data was injected at the 393-second mark.
\begin{figure}[t]
\scriptsize
	\centering
	\includegraphics[width=0.8\columnwidth]{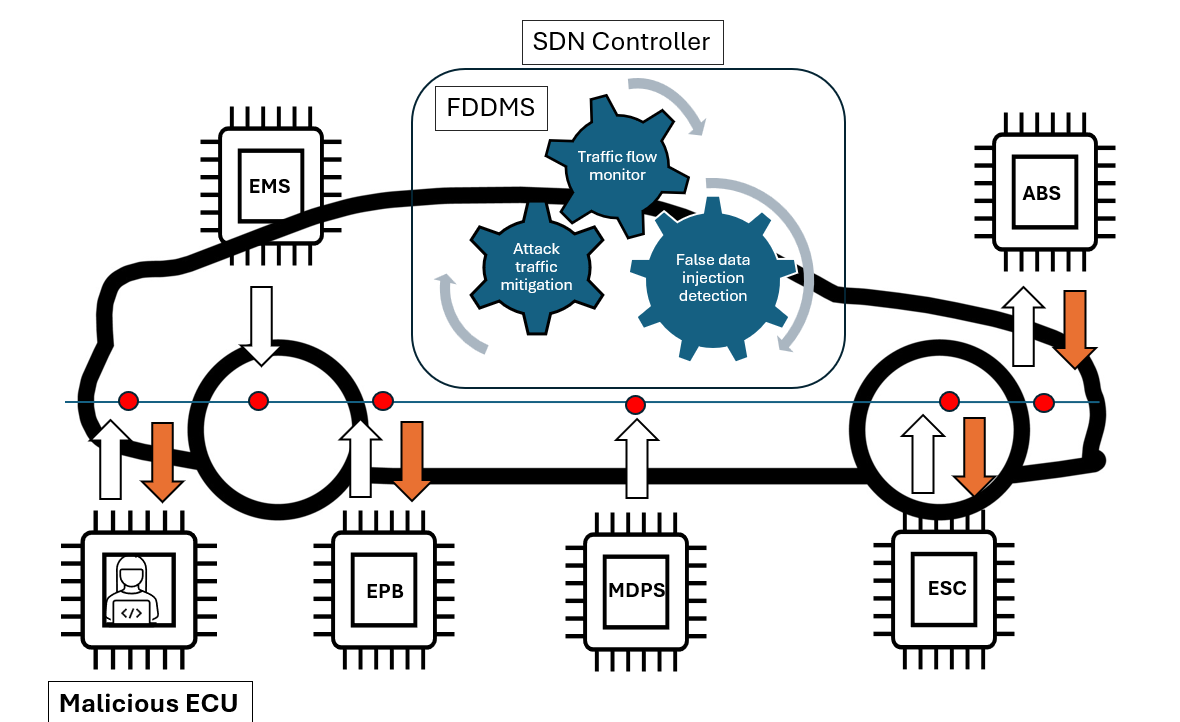}  
	\caption{SDN-based In-vehicle network}
	\label{fig:framework}
\end{figure}

\subsection{Generating adversarial examples}
\subsubsection{Fast gradient sign method (FGSM)}
\authone{Goodfellow et al.~\cite{GoodfellowFGSM} introduced FGSM for generating adversarial examples against deep learning models. FGSM crafts adversarial examples $x^\prime$ by adding perturbations in the direction of gradient of the loss function, that is,\vspace{-1.8mm}
\begin{equation}
    x^\prime = x + \epsilon \times sign\left( (gradient_{x}{L}(F(x),y; \theta)) \right),
\end{equation}
where $\epsilon$ controls the perturbation. To reduce distortion, a smaller $\epsilon$ should be used, but this could also decrease the success rate of evading the detector.}
\subsubsection{Basic Iterative Method (BIM):}
\authone{Kurakin et al.~\cite{DBLP:journals/corr/KurakinGB16} proposed an iterative version of FGSM called BIM by taking multiple small steps in the gradient direction. At each iteration $t$, the point $x^t$ was updated by:\vspace{-1.8mm}
\begin{equation}
\label{BIM}
    x^{t+1} = Clip_{\epsilon} (x^t + \alpha \times sign( gradient_{x}(L(F(x^t),y; \theta)))),
\end{equation}
where $\alpha$ is the step size for each iteration. \authtwo{Kurakin et al.~\cite{DBLP:journals/corr/KurakinGB16} obtained adversarial examples $x^{t+1}$ that could lead to misclassifications after a number of steps, $m$.} By using a small step size $\alpha = \frac{\epsilon}{m}$ in each step, BIM might find more subtle and adversarial perturbations compared to a single large step in FGSM. Similarly to FGSM, the parameter $\epsilon$ should be tuned to minimize distortion while maintaining a high ASR. In our experiment, we set the initial point $x^{0}$ to the original example $x$.}
\subsubsection{DeepFool}
\authone{Given the FDIA signals, we used the DeepFool attack to obtain the minimal perturbation needed to bypass the detection model. In a binary classification problem, DeepFool~\cite{Moosavi-Dezfooli16} linearly approximated the decision boundary of a machine learning model around an input point. The minimal perturbation was obtained by calculating the distance between the input point and its projection on the decision boundary, then plus an overshoot amount of $\epsilon = 0.02$ to ensure misclassifications. To speed up the noise generation process, in each iteration, the attack modified the intermediary signals $x'$ by $r_{tot}$, which is an accumulation of ongoing perturbations $r_i$. DeepFool continued until the model misclassified $x'$ or a maximum iteration $T$ was reached.} 

\authone{We further made two modification to the DeepFool~\cite{Moosavi-Dezfooli16} as shown in Algorithm~\ref{algo:deepfool} to enhance its effectiveness for the FDIA signals: 
\begin{itemize}
    \item Gradient Clipping: because the gradient w.r.t $x'$ can suffer from the gradient explosive problem, especially for recurrent neural networks, the perturbation $r_i$ can be large. Thus, we limited the gradient as follows:
\begin{equation}
Gradient Clipping(x) = 
        \begin{cases}
            min\_value & \text{ if } x < min\_val \\
            max\_value & \text{ if } x > max\_val \\
            x & \text{otherwise} \\
        \end{cases}
\end{equation} In our experiments, we set the $min\_val = max\_val = \alpha$ where $\alpha = 0.95$. 
\end{itemize}
\begin{itemize}
    \item Scaling Factor: we further reduced the perturbation $r_i$ by multiplying it by the scaling factor $\kappa$ due to the accumulation nature of our update rule. Specifically, we updated $r_{tot}$ as follows, \vspace{-1.8mm}
    \begin{equation}
        r_{tot} = r_{tot} + \kappa \times r_i
    \end{equation} where $\kappa$ is less than 1. We used $\kappa = {0.5}$ for our experiments. 
\end{itemize}} \vspace{-1.8mm}

\begin{algorithm}
\caption{The DeepFool variant for detection models}
\begin{algorithmic}
\STATE \textbf{input:} $x$: an input signal, $model$: a detection model, $T$ the maximum iterations, $\epsilon$: the overshooting factor, $\kappa$: the step decaying factor, $\alpha$: the gradient clipping factor \\
\STATE \textbf{output:} $\hat{r}$: a minimum distorted perturbation and $x'$: a perturbed signal.
\STATE Initialize $t \leftarrow 0$, $step \leftarrow \epsilon$, $r_{tot} \leftarrow 0$
\WHILE{$t < T$}
    \STATE $x_i \leftarrow x + r_{tot}$
    \STATE $loss \leftarrow model(x_i)$
    \STATE $grad \leftarrow gradient_{x_i}(loss)$
    \STATE $grad \leftarrow Gradient Clipping_{[-\alpha, +\alpha]}(grad)$
    \STATE $r_i \leftarrow \frac{loss \cdot grad}{||grad||_2^2}$
    \STATE $r_{tot} \leftarrow r_{tot} + \kappa \times r_i$
    \IF{$pred(x) != pred(x_i)$}
        \RETURN $r_{tot}$ \hfill // The perturbation is minimized.
    \ENDIF
    \STATE $t \leftarrow t + 1$
\ENDWHILE
\RETURN $\hat{r} \leftarrow r_{tot}$, $\hat{x} \leftarrow x + (1 + \epsilon) \times r_{tot}$
\end{algorithmic}
\label{algo:deepfool}
\end{algorithm}

\subsection{An SDN-based FDDMS For In-vehicle Networks}
In an in-vehicle network, the CAN bus is characterized by a security challenge where multiple ECUs broadcast and receive CAN messages without explicit source and destination addresses. Securing these communications is essential for ensuring the vehicle's safety and security. Software-Defined Networking (SDN) architecture provides a centralized controller with a global view of the entire in-vehicle network. The SDN controller can be programmed to implement and deploy sophisticated security mechanisms to manage the network. This global network view and programmability enable the SDN controller to monitor and manage network traffic in real-time, providing improved oversight and control.
Implementing SDN in an in-vehicle network improves communication flow management while enhancing vehicle security and safety.

In this study, we designed and implemented an SDN-based FDDMS to detect and mitigate FDIA in in-vehicle networks. We simulated a malicious ECU
that is connected to the CAN bus. Figure~\ref{fig:framework} illustrates the overview of the proposed SDN-based in-vehicle network, which comprises six ECUs connected to the CAN bus. EMS and MDPS ECUs are restricted to broadcasting CAN messages, whereas ABS, ESC, and EPB ECUs are designed to receive them. In contrast, a compromised ECU can broadcast and receive messages on the CAN bus.
%
We placed an OpenFlow Switch (OVS) on the CAN bus and connected it to the SDN controller. This placement allows the controller to manage and monitor the traffic flow through the CAN bus via the OVS. 
%
During normal operation, the EMS ECU broadcasts legitimate EMS11, EMS12, EMS14, and EMS16 CAN messages, while the MDPS ECU transmits attack-free SAS11 CAN messages over the CAN bus. However, the compromised ECU, after intercepting the messages exchanged on the bus, can alter some chosen messages by injecting false data. These tampered messages are then broadcast onto the CAN bus, where ECUs such as ABS, ESC, and EPB receive the altered messages and may take actions based on the manipulated data.


The FDDMS framework was implemented in the SDN controller, which has a global view of the network. This global network view allows the controller to monitor all traffic flows going through the network. The framework consisted of three main phases. Initially, the SDN controller continuously monitors the traffic flowing through the CAN bus. Next, an LSTM-based detection model identifies FDIA. Finally, the mitigation module redirects the attack traffic to a backend storage system. Specifically, if an attack was detected, the SDN controller would update the flow table entries in the OVS, altering the network's forwarding rules. As a result, attack messages are no longer broadcast over the CAN bus but are instead sent to the backend storage for further analysis.

\subsubsection{\textbf{The LSTM based Detection Model}}
In the FDDMS framework, we developed an LSTM-based detection model to identify FDIA. LSTM excels at processing time-series data like network traffic or time-series data from in-vehicle network~\cite{li2021adversarial}. In our study, once the raw data was decoded, we selected the 20 features associated with the five brake related ECUs as input features for the LSTM model. Because LSTM models are designed for sequential data, we re-organized the time-series data of the selected signals into fixed length sequences. We fed 20 features into the LSTM model, each feature being a sequence of decoded CAN signal values.

In our model, the LSTM layer consisted of 128 neurons. The LSTM layer is best known for their recurrent structure, where each element of a sequence is processed in the same manner, with the output being dependent on previous calculations. LSTM networks were introduced to address issues like vanishing gradients and to capture long-term dependencies by introducing self-loops, allowing information to persist over time~\cite{Goodfellow-et-al-2016}.

An LSTM typically contains three gates: an input gate, an forget gate, and an output gate. After the LSTM layer, the output layer consisted of a single neuron, where the output is 0 (Normal) or 1 (Attack). The output layer employs a sigmoid function, which converts the output to a probability, making it suitable for the binary classification task. For the loss function, we used binary cross-entropy, which is ideal for binary classification. To optimize the model, we employed the Adam optimizer~\cite{kingma2014adam}. Additionally, in the Evaluation section, we will demonstrate that the LSTM model can be effectively optimized using RMSProp, Adagrad, and Adam.

Our dataset consists of 1,894 instances in total. We splitted the data as follows: 80\% for training, 10\% for validation, and 10\% for testing. Specifically, this means that we used 1,516 samples for training, 189 samples for validation, and 189 samples for testing.

\subsubsection{\textbf{Adversarial Training with The Sample Selection Technique}}
\authone{To overcome the destructive impact caused by the adversarial attacks, we adopted an AdvTrain technique~\cite{li2021adversarial} that effectively improved the detection model's robustness against FGSM and BIM. Compared to~\cite{MadryMSTV17}, Li.et al's re-training technique~\cite{li2021adversarial} gradually increased the training set size and re-traind the detection model using both adversarial examples and original training examples in each iteration. We further extended the re-training technique by proposing a score base strategy to only select challenging adversarial examples for training. Let $Selection(S_{k})$ denote a decision on whether to select the sample $S_{k}$ with its ground truth label $y_k$} \vspace{-1.8mm}
\begin{equation}
\label{eq:threshold_select}
    Selection(S_{k})= \begin{cases}
        Yes & \text{if } score < threshold\\
        No & \text{if } otherwise,
    \end{cases}
\end{equation}
Then, the score can be computed using a function $g(.)$,
\begin{equation}
\label{eq:score_func}
    score = g(S_{k}, y_k)
\end{equation} \authone{where $threshold$ is a parameter. Using the $threshold = 0.5$ and $g(S_{k}, y_k)$ based on the LSTM model's prediction w.r.t $y_k$, we computed the scores for each sample $S_{k}$. We only trained the model on those examples with a score below the threshold. Thus, we forced the model to learn the patterns given by those challenging examples instead of easy ones. Our proposed AdvTrain with the sample selection technique is shown in Algorithm~\ref{algo:at}. To disrupt the carefully crafted perturbation by adversarial attacks, Lin, et al~\cite{lin2020secure} added Gaussian noise to test samples. Different from~\cite{lin2020secure}, we added adversarial noises generated by FGSM and BIM to test samples.} \vspace{-1.8mm}

\begin{algorithm}
\begin{algorithmic}[1]
\STATE \textbf{Input} $R$: number of iterations; $S^{repo}$: the adversarial example repository; $score$: the score function to select samples for expanding $S^{repo}$; $N$: number of randomly chosen sample; $\theta^{0}$: the pretrained model weights.
\STATE \textbf{Output}: The LSTM model that is robust to adversarial attacks. 
    \FORALL{ $t = 1$ \textbf{to} $R$}
        \STATE Randomly choose $N$ instances $S_t$ from the training dataset;
        \FORALL{$k = 1$ \textbf{to} $N$}
            \STATE Use adversarial model to generate $S_{t,k}^{'}$;
            \STATE Calculate a score based on Equation~\ref{eq:score_func} for each adversarial instance $S_{t,k}^{'}$;
            \STATE Use the selection criteria in Equation~\ref{eq:threshold_select} to add $S_{t, k}^{'}$ to $S^{repo}$
        \ENDFOR
        \STATE Randomly choose $N$ adversarial examples $S_t^{adv}$;
        \STATE Update $\theta^{t}$ by training on $S_t + S_t^{adv}$;
    \ENDFOR
\end{algorithmic}
\caption{AdvTrain with the sample selection technique} 
\label{algo:at}
\end{algorithm}

\subsubsection{\textbf{The Mitigation Scheme}}
In an SDN network, when an attack is detected, the SDN controller can modify the flow table rules within the OpenFlow Switch (OVS). The controller and the OVS interact through a secure channel implemented by a protocol called OpenFlow. Each flow table entry comprises three essential fields: (a) a header field that identifies the flow, (b) an action field that instructs on how the traffic flow should be redirected, and (c) a statistics field that records network data related to the flow.

In this study, we utilized the three characteristics of SDN network to update flow table rules, preventing false messages from being broadcasted on the CAN bus. Instead, these messages were redirected to backend storage. When the LSTM-based detection model identified attack traffic, the mitigation process was triggered.
In the mitigation scheme, an OVS flow rule was applied to redirect attack traffic to backend storage rather than allowing it to reach the CAN bus, while simultaneously sending an alert to the driver. If no attack was detected, the messages were broadcast normally.
\section{Experimental Results}\label{sec:evaluation}
The SDN-based in-vehicle network was set up and tested on GENI~\cite{GENI-Berman}, using Floodlight as the SDN controller~\cite{floodlight}. The network comprises six ECUs: EMS, MDPS, ESC, EPB, ABS, and a malicious ECU. The EMS and MDPS broadcast legitimate CAN messages, while the malicious ECU injects false messages. The ESC, EPB, and ABS ECUs were configured to only receive messages. This section first evaluates the LSTM-based detection model with normal and robust accuracy, followed by an evaluation on the proposed re-training technique. Lastly, the section concludes with an investigation on the computational complexity of the SDN-based mitigation scheme.

\subsection{Evaluation of the LSTM model}
\subsubsection{Normal accuracy}
In the LSTM-based detection model, the LSTM layer consisted of 128 neurons. We set the number of epochs to 50. We measured the model performance using the "normal" accuracy and time. The "Time" column shows the computational costs required to train the model, measured in seconds. To ensure a comprehensive evaluation of our proposed model, we trained our proposed model using multiple optimization algorithms, including SGD, Adam, RMSprop, and Adagrad. Specifically, we set SGD, RMSprop, and Adam with a learning rate of 0.001, while Adagrad's learning rate was set to 0.1. Table~\ref{tab:eval} shows the normal accuracy, recall, precision, F1 score, and training time for each optimizer. We conducted the experiments in GAIVI cluster at the University of South Florida and used Pytorch to create and train the LSTM model.\vspace{-1.8mm}
\begin{table}[h]
    \caption{Evaluation of Different Optimizers}
    \small\addtolength{\tabcolsep}{-1.8pt}
    \begin{tabular}{| l | l | l | l | l | l |} 
    \hline
     {\textbf{Optimizer}} &{\multirow{2}{1.4cm}{\textbf{Normal \\Accuracy}}} & {\textbf{Recall}}  & {\textbf{Precision}} & {\textbf{F1 Score}} & {\textbf{Time(s)}} \\
      & & & & &  \\ 
      \hline
      RMSprop & 99.47\% & 1.00 & 0.99 & 1.00 & 48.27\\ 
      \hline
      Adam    & 98.95\% & 0.99 & 0.99 & 0.99 & 56.60\\ 
      \hline
      Adagrad & 99.47\% & 1.00 & 0.99 & 1.00 & 93.94\\ 
      \hline
      SGD     & 54.21\% & 0.33 & 0.62 & 0.43 & 48.13\\ 
      \hline
    \end{tabular}
    \label{tab:eval}
\end{table}

As shown in Table~\ref{tab:eval}, the RMSprop and Adagrad optimizer achieved the highest normal accuracy of 99.47\%. Their recall, precision, and F1 score were 1.00, 0.99, and 1.00, respectively. The Adam optimizer obtained the second best performance with a normal accuracy of 98.95\%. Its recall, precision, and F1 score were all equal to 0.99. The training times for the RMSprop, Adam, Adagrad, and SGD optimizers were 48.27 seconds, 56.60 seconds, 93.94 seconds, and 48.13 seconds, respectively. The training times are notably short, allowing the FDDMS to update the model every two minutes and perform real-time detection.
\subsubsection{Normal accuracy, robust accuracy, and adversarial training}
Nowadays, it is crucial to measure the model performance using both the normal and robust accuracy. We followed the setup detailed in Li, et al,~\cite{li2021adversarial} to evaluate the robustness of our models. We set the batch size N = 200, and in each iteration, we re-trained the model for 30 epochs. We adversarially trained the models using the Adam optimizer with a learning rate of 0.001. At test time, in addition to augmenting the training dataset with the FGSM and BIM examples, we added FGSM and BIM perturbations to test examples to improve the model's detection ability~\cite{lin2020secure}. We computed the normal accuracy using the test examples with FDIA. We used FGSM-$L_{2}$, BIM-$L_{2}$, the original DeepFool, and our extended DeepFool to calculate the robust accuracy. Furthermore, we used three different datasets for training: 1) the FDIA based, 2) the FGSM-$L_{2}$, and BIM-based-$L_{2}$ datasets. \authone{Details about how to generate the four types of adversarial examples were discussed in the Methodology section}. The results of the experiment are presented in Table~\ref{table:AT_robustness}, where our adversarially trained models achieved the best normal and robust accuracy.

The LSTM based model trained on the challenging FGSM examples achieved the best normal accuracy of 99.47\% ,and the robust accuracy against FGSM, BIM, the original DeepFool, and our DeepFool($\kappa=0.5, \alpha=0.95$)  is 99.475\%, 99.47\%, 98.95\%, and 99.47\%, respectively. When we adversarially trained the model using the challenging BIM examples, we achieved the same normal accuracy and the same robust accuracy for FGSM and BIM. However, the robust accuracy against the DeepFool attacks decreased to 93.16\% and 94.21\%. In contrast, if we trained with only clean FDIA signal, the robust accuracy was significantly reduced to 37\%, 32\%, 1\%, and 1\% for FGSM, BIM, the original DeepFool, and our DeepFool attacks, respectively.\vspace{-1.8mm} 



\begin{table}[h]
    \centering
    \small\addtolength{\tabcolsep}{-1.8pt}
    \footnotesize
    \caption{Normal accuracy and robust accuracy (\%) of the LSTM based detection model under various white-box attacks.}
    \begin{tabular}{| l | l | l | l | l | l |}
    \hline
        \multirow{2}{1.0cm}{Training \\set} & FDIA & FGSM-$L_2$ & BIM-$L_2$ & DeepFool & \multirow{2}{1.0cm}{Our \\DeepFool} \\ 
        & & & & & \\
        \hline
        FDIA & 100 & 37 & 32 & 1 & 1  \\ \hline
        FGSM & 97.89 & 99.47 & 96.32 & 63.68 & 65.26 \\ \hline
        Challenging FGSM & 99.47 & 99.475 & 99.47 & 98.95 & 99.47 \\ \hline
        BIM & 99.47 & 97.89 & 99.47 & 25.26 & 14.74 \\ \hline
        Challenging BIM & 99.47 & 99.47 & 99.47 & 93.16 & 94.21 \\ \hline
    \end{tabular}
    \label{table:AT_robustness}
\end{table}
\subsection{Evaluation on the impacts of adversarial attacks}
In this section, we compare FGSM, BIM, the original DeepFool, and our extended DeepFool in terms of the perturbation magnitude. To ensure a comprehensive evaluation of the perturbation magnitude, we adopt the mean $L_{0}$, $L_{2}$, $L_{\infty}$ to measure the distance between the original signals and the perturbed ones. We only selected the attack signals from the test set to perform adversarial attacks.
The $L_{0}$ metric counts the number of perturbed features. The $L_{2}$ metric reports the Euclidean distance between the original and perturbed signals, while the $L_{\infty}$ measures the maximum absolute change applied to a feature. Table~\ref{tab:attack_comparison} presents the attack success rate (\%) and the mean distortion metrics for each method. 
The hyperparameters for each attack method are listed in Table~\ref{tab:param_summary}. We generated adversarial examples for FGSM and BIM via the IBM Adversarial Machine Learning toolbox~\cite{art2018}. 
Table~\ref{tab:attack_comparison} shows that our extended DeepFool achieved the lowest $L_{\infty}$-norm, $L_{0}$, and $L_{2}$-norm when using ($\kappa=0.5, \alpha=0.95$), indicating that it introduces minimal perturbation while maintaining comparable high ASR of 99\% when compared to FGSM$-L_{\infty}$, BIM$-L_{\infty}$, and the original DeepFool ($L_{2}$) attacks, respectively. The use of gradient clipping and scaling factors effectively reduced perturbation while maintaining a high ASR, making it an effective method to make the FDIA signal robust against the detection model.
\begin{table}[h!]
    \caption{Adversarial attack Hyperparameter summary}
    \centering
    \small\addtolength{\tabcolsep}{-1.0pt}
    \footnotesize
    \label{tab:param_summary}
    \begin{tabular}{| l | l | l | l | l | l |}
    \hline
    {\textbf{Attack Method}} & {\textbf{$\epsilon$}} & {\textbf{Step Size}} & {\textbf{Iterations}} & {\textbf{$\kappa$}} & {\textbf{$\alpha$}}  \\ \hline
    FGSM$-L_{2}$                & 13.0  & - & 1 & - & -  \\ \hline
    FGSM$-L_{\infty}$           & 13.0  & - & 1 & - & -  \\ \hline
    BIM$-L_{2}$                 & 13.0  & 13.0/20 & 20 & - & - \\ \hline
    BIM$-L_{\infty}$            & 13.0  & 13.0/20 & 20 & - & - \\ \hline
    DeepFool ($L_{2}$)          & -     & 0.02 & 50 & - & -     \\ \hline
    Our DeepFool ($L_{2}$)      & -     & 0.02 & 50 & 0.5 & 0.95 \\ \hline
\end{tabular}
\end{table}
\begin{table}[h!]
    \caption{Comparison of different adversarial attacks on the CAN data}
    \small\addtolength{\tabcolsep}{-1.0pt}
    \footnotesize
    \label{tab:attack_comparison}
    \begin{tabular}{| l | l | l | l | l |}
    \hline
    {\textbf{Attack Method}} & {\textbf{ASR}} & {\textbf{$L_{0}$}} & {\textbf{$L_{2}$}} & {\textbf{$L_{\infty}$}} \\ \hline
    FGSM$-L_{2}$                & 63.00              & 4337.56             & 67.32             & 4.85                     \\ \hline
    FGSM$-L_{\infty}$       & 99.00              & 17198.69             & 2199559.0             & 13.00                     \\ \hline
    BIM$-L_{2}$                 & 68.00              & 2705.38            & 77.89             & 5.04                     \\ \hline
    BIM$-L_{\infty}$         & 99.00              & 17952.02             & 1450620.5             & 13.00                     \\ \hline
    DeepFool ($L_{2}$)          & 99.00             & 6647.62             & 53.66            & 3.54                     \\ 
    \hline
    {DeepFool ($\kappa=0.5, \alpha=0.95$)} & 99.00              & \textbf{6543.51} & \textbf{38.65} & \textbf{3.00 }                    \\ 
    \hline
\end{tabular}
\end{table}

\subsection{Evaluation of the SDN and Mitigation Scheme}
In this section, we evaluate the transmission latency of our SDN-based in-vehicle network, as well as the detection and mitigation time of our FDDMS. Because the interval between ECU are typically broadcast every 10 ms in this dataset, an overall latency below this threshold demonstrates that a security system does not disrupt normal vehicle operation. 
%
Table~\ref{tab:time} presents the latency for each message type, including detection time, the mitigation time, and the total latency. The overall latency is 7.7629 ms. The overall latency below the 10 ms threshold shows that our FDDMS can operate effectively without disrupting normal vehicle operation.\vspace{-1.8mm}
%
\begin{table}[h]
    \caption{Evaluation of Mitigation Time}
    \small\addtolength{\tabcolsep}{-1.8pt} 
    \begin{tabular}{| c | c | c | c | c |}  
    \hline
     {\textbf{Msg Type}} &
     {\multirow{2}{1.0cm}{\textbf{Latency (ms)}}} & {\multirow{2}{1.5cm}{\textbf{Detection \\Time (ms)}}}  & {\multirow{2}{1.5cm}{\textbf{Mitigation Time (ms)}}}  & {\multirow{2}{2cm}{\textbf{Overall \\Latency (ms)}}} \\ 
      &&&& \\ \hline
      EMS11  & 5.5490 & \multirow{7}{1.5cm}{\hspace{10pt} 1.6021} & \multirow{7}{1.5cm}{\hspace{10pt} 0.7991} &  7.9502\\ \cline{1-2} \cline{5-5}
      EMS12   & 5.7007 & & & 8.1019\\ \cline{1-2} \cline{5-5}
      EMS14   & 5.1252 & & & 7.5264\\ \cline{1-2} \cline{5-5}
      EMS16   & 5.0749 & & & 7.4761\\ \cline{1-2} \cline{5-5}
      SAS11   & 4.9052 & & & 7.3064\\ \cline{1-2} \cline{5-5}
      Attack  & 5.8154 & & & 8.2166\\ \cline{1-2} \cline{5-5}
      Average & 5.3617 & & & 7.7629\\ \hline
    \end{tabular}
    \label{tab:time}
\end{table}
\vspace{-1.8mm}
\section{Conclusion}\label{sec:conclusion}



This research proposed an SDN-based False Data Detection and Mitigation System (FDDMS) for in-vehicle networks, utilizing a real-world KIA SOUL CAN dataset. The raw data was decoded using an OpenDBC file, and an attack model was developed to simulate false data injection into the CAN bus. The system incorporates five brake-related ECUs and a malicious ECU for injecting attack traffic. The SDN controller continuously monitors network traffic, while the FDDMS, which includes an LSTM-based detection model and a mitigation strategy, identifies and redirects malicious data. Our approach achieved a detection accuracy of 99.47\%. We also develop an effective variant of the DeepFool attack to evaluate the robustness of the detection mode. To countermeasure the extended DeepFool attack, we further enhance the LSTM's robustness with an adversarial re-training approach. The re-training approach selects strong adversarial examples for training. Future work will focus on developing more sophisticated attack models and further enhancing the FDDMS to improve defenses against adversarial attacks.
\section*{Acknowledgment}
We acknowledge NSF for partially sponsoring the work under grants \#1620868 with its REU, \#2228562, and \#2236283. We also thank Cyber Florida for a seed grant.

\bibliographystyle{IEEEtran}
\bibliography{reference}

\begin{thebibliography}{10}
\providecommand{\url}[1]{#1}
\csname url@samestyle\endcsname
\providecommand{\newblock}{\relax}
\providecommand{\bibinfo}[2]{#2}
\providecommand{\BIBentrySTDinterwordspacing}{\spaceskip=0pt\relax}
\providecommand{\BIBentryALTinterwordstretchfactor}{4}
\providecommand{\BIBentryALTinterwordspacing}{\spaceskip=\fontdimen2\font plus
\BIBentryALTinterwordstretchfactor\fontdimen3\font minus \fontdimen4\font\relax}
\providecommand{\BIBforeignlanguage}[2]{{%
\expandafter\ifx\csname l@#1\endcsname\relax
\typeout{** WARNING: IEEEtran.bst: No hyphenation pattern has been}%
\typeout{** loaded for the language `#1'. Using the pattern for}%
\typeout{** the default language instead.}%
\else
\language=\csname l@#1\endcsname
\fi
#2}}
\providecommand{\BIBdecl}{\relax}
\BIBdecl

\bibitem{li2020countermeasures}
Y.~Li, ``Countermeasures against various network attacks using machine learning methods,'' Doctoral dissertation, University of South Florida, 2020.

\bibitem{mudalige2015efficient}
U.~P. Mudalige and M.~Losh, ``Efficient data flow algorithms for autonomous lane changing, passing and overtaking behaviors,'' Aug.~4 2015, uS Patent 9,096,267.

\bibitem{wang2019survey}
J.~Wang, Y.~Shao, Y.~Ge, and R.~Yu, ``A survey of vehicle to everything (v2x) testing,'' in \emph{Sensors}, vol.~19, no.~2, 2019, p. 334.

\bibitem{huang2018vehicle}
J.~Huang, M.~Zhao, Y.~Zhou, and C.-C. Xing, ``In-vehicle networking: Protocols, challenges, and solutions,'' in \emph{IEEE Network}, vol.~33, no.~1.\hskip 1em plus 0.5em minus 0.4em\relax IEEE, 2018, pp. 92--98.

\bibitem{avatefipour2018state}
O.~Avatefipour and H.~Malik, ``State-of-the-art survey on in-vehicle network communication \mbox{(CAN-Bus)} security and vulnerabilities,'' \emph{arXiv:1802.01725}, 2018.

\bibitem{carsten2015vehicle}
P.~Carsten, T.~R. Andel, M.~Yampolskiy, and J.~T. McDonald, ``In-vehicle networks: Attacks, vulnerabilities, and proposed solutions,'' in \emph{CISRC}.\hskip 1em plus 0.5em minus 0.4em\relax ACM, 2015, p.~1.

\bibitem{kreutz2015software}
D.~Kreutz, F.~M. Ramos, P.~E. Verissimo, C.~E. Rothenberg, S.~Azodolmolky, and S.~Uhlig, ``Software-defined networking: A comprehensive survey,'' in \emph{Proceedings of the IEEE}, vol. 103, no.~1.\hskip 1em plus 0.5em minus 0.4em\relax IEEE, 2015, pp. 14--76.

\bibitem{islam2018cybersecurity}
M.~Islam, M.~Chowdhury, H.~Li, and H.~Hu, ``Cybersecurity attacks in vehicle-to-infrastructure applications and their prevention,'' in \emph{Transportation Research Record}, vol. 2672, no.~19.\hskip 1em plus 0.5em minus 0.4em\relax SAGE, 2018, pp. 66--78.

\bibitem{liang20162015}
G.~Liang, S.~R. Weller, J.~Zhao, F.~Luo, and Z.~Y. Dong, ``The 2015 ukraine blackout: Implications for false data injection attacks,'' in \emph{IEEE Transactions on Power Systems}, vol.~32, no.~4.\hskip 1em plus 0.5em minus 0.4em\relax IEEE, 2016, pp. 3317--3318.

\bibitem{DBLP:journals/corr/abs-2112-02797}
J.~Lin, L.~Dang, M.~Rahouti, and K.~Xiong, ``{ML} attack models: Adversarial attacks and data poisoning attacks,'' \emph{arXiv:2112.02797}, 2021.

\bibitem{GoodfellowFGSM}
I.~J. Goodfellow, J.~Shlens, and C.~Szegedy, ``Explaining and harnessing adversarial examples,'' in \emph{ICLR}, 2015.

\bibitem{DBLP:journals/corr/KurakinGB16}
A.~Kurakin, I.~J. Goodfellow, and S.~Bengio, ``Adversarial examples in the physical world,'' \emph{arXiv:1607.02533}, 2016.

\bibitem{Moosavi-Dezfooli16}
S.~Moosavi{-}Dezfooli, A.~Fawzi, and P.~Frossard, ``Deepfool: {A} simple and accurate method to fool deep neural networks,'' in \emph{CVPR}, 2016, pp. 2574--2582.

\bibitem{cerracchio2024investigating}
P.~Cerracchio, S.~Longari, M.~Carminati, S.~Zanero \emph{et~al.}, ``Investigating the impact of evasion attacks against automotive intrusion detection systems,'' in \emph{Symposium on Vehicle Security and Privacy}, 2024.

\bibitem{MadryMSTV17}
A.~Madry, A.~Makelov, L.~Schmidt, D.~Tsipras, and A.~Vladu, ``Towards deep learning models resistant to adversarial attacks,'' \emph{arXiv:1706.06083}, 2017.

\bibitem{ilyas2019adversarial}
A.~Ilyas, S.~Santurkar, D.~Tsipras, L.~Engstrom, B.~Tran, and A.~Madry, ``Adversarial examples are not bugs, they are features,'' in \emph{NeurIPS}, vol.~32, 2019.

\bibitem{athalye2018obfuscated}
A.~Athalye, N.~Carlini, and D.~Wagner, ``Obfuscated gradients give a false sense of security: Circumventing defenses to adversarial examples,'' in \emph{ICML}, 2018, pp. 274--283.

\bibitem{lin2020secure}
J.~Lin, L.~L. Njilla, and K.~Xiong, ``Robust machine learning against adversarial samples at test time,'' in \emph{IEEE International Conference on Communications}, 2020, pp. 1--6.

\bibitem{xie2020smooth}
C.~Xie, M.~Tan, B.~Gong, A.~Yuille, and Q.~V. Le, ``Smooth adversarial training,'' \emph{arXiv:2006.14536}, 2020.

\bibitem{zhang2019theoretically}
H.~Zhang, Y.~Yu, J.~Jiao, E.~Xing, L.~El~Ghaoui, and M.~Jordan, ``Theoretically principled trade-off between robustness and accuracy,'' in \emph{ICML}, 2019, pp. 7472--7482.

\bibitem{li2021adversarial}
Y.~Li, J.~Lin, and K.~Xiong, ``An adversarial attack defending system for securing in-vehicle networks,'' in \emph{IEEE Consumer Communications \& Networking Conference}, 2021, pp. 1--6.

\bibitem{lee2017otids}
H.~Lee, S.~H. Jeong, and H.~K. Kim, ``Otids: A novel intrusion detection system for in-vehicle network by using remote frame,'' in \emph{PST}.\hskip 1em plus 0.5em minus 0.4em\relax IEEE, 2017, pp. 57--5709.

\bibitem{liu2017vehicle}
J.~Liu, S.~Zhang, W.~Sun, and Y.~Shi, ``In-vehicle network attacks and countermeasures: Challenges and future directions,'' in \emph{IEEE Network}, vol.~31, no.~5.\hskip 1em plus 0.5em minus 0.4em\relax IEEE, 2017, pp. 50--58.

\bibitem{pham2021survey}
M.~Pham and K.~Xiong, ``A survey on security attacks and defense techniques for connected and autonomous vehicles,'' \emph{Computers \& Security}, vol. 109, p. 102269, 2021.

\bibitem{muter2010structured}
M.~M{\"u}ter, A.~Groll, and F.~C. Freiling, ``A structured approach to anomaly detection for in-vehicle networks,'' in \emph{IAS}.\hskip 1em plus 0.5em minus 0.4em\relax IEEE, 2010, pp. 92--98.

\bibitem{muter2011entropy}
M.~M{\"u}ter and N.~Asaj, ``Entropy-based anomaly detection for in-vehicle networks,'' in \emph{IEEE IV}.\hskip 1em plus 0.5em minus 0.4em\relax IEEE, 2011, pp. 1110--1115.

\bibitem{zhao2024potential}
X.~Zhao, Y.~Fang, H.~Min, X.~Wu, W.~Wang, and R.~Teixeira, ``Potential sources of sensor data anomalies for autonomous vehicles: An overview from road vehicle safety perspective,'' \emph{Expert Systems with Applications}, vol. 236, p. 121358, 2024.

\bibitem{baccari2024anomaly}
S.~Baccari, M.~Hadded, H.~Ghazzai, H.~Touati, and M.~Elhadef, ``Anomaly detection in connected and autonomous vehicles: A survey, analysis, and research challenges,'' \emph{IEEE Access}, 2024.

\bibitem{fang2024anomaly}
Y.~Fang, H.~Min, X.~Wu, W.~Wang, X.~Zhao, B.~Martinez-Pastor, and R.~Teixeira, ``Anomaly diagnosis of connected autonomous vehicles: A survey,'' \emph{Information Fusion}, vol. 105, p. 102223, 2024.

\bibitem{cho2016fingerprinting}
K.-T. Cho and K.~G. Shin, ``Fingerprinting electronic control units for vehicle intrusion detection,'' in \emph{$USENIX$ Security Symposium}, 2016, pp. 911--927.

\bibitem{kang2016intrusion}
M.-J. Kang and J.-W. Kang, ``Intrusion detection system using deep neural network for in-vehicle network security,'' in \emph{PloS one}, vol.~11, no.~6, 2016, p. e0155781.

\bibitem{kuwahara2018supervised}
T.~Kuwahara, Y.~Baba, H.~Kashima, T.~Kishikawa, J.~Tsurumi, T.~Haga, Y.~Ujiie, T.~Sasaki, and H.~Matsushima, ``Supervised and unsupervised intrusion detection based on can message frequencies for in-vehicle network,'' in \emph{Journal of Information Processing}, vol.~26, 2018, pp. 306--313.

\bibitem{liu2011false}
Y.~Liu, P.~Ning, and M.~K. Reiter, ``False data injection attacks against state estimation in electric power grids,'' in \emph{ACM TISSEC}, vol.~14, no.~1.\hskip 1em plus 0.5em minus 0.4em\relax ACM, 2011, p.~13.

\bibitem{guan2015comprehensive}
Z.~Guan, N.~Sun, Y.~Xu, and T.~Yang, ``A comprehensive survey of false data injection in smart grid,'' in \emph{International Journal of Wireless and Mobile Computing}, vol.~8, no.~1.\hskip 1em plus 0.5em minus 0.4em\relax Inderscience Publishers, 2015, pp. 27--33.

\bibitem{jeba2012false}
S.~A. Jeba and B.~Paramasivan, ``False data injection attack and its countermeasures in wireless sensor networks,'' in \emph{European Journal of Scientific Research}, vol.~82, no.~2, 2012, pp. 248--257.

\bibitem{cao2008proof}
Z.~Cao, J.~Kong, U.~Lee, M.~Gerla, and Z.~Chen, ``Proof-of-relevance: Filtering false data via authentic consensus in vehicle ad-hoc networks,'' in \emph{IEEE INFOCOM Workshops}.\hskip 1em plus 0.5em minus 0.4em\relax IEEE, 2008, pp. 1--6.

\bibitem{moore2017modeling}
M.~R. Moore, R.~A. Bridges, F.~L. Combs, M.~S. Starr, and S.~J. Prowell, ``Modeling inter-signal arrival times for accurate detection of can bus signal injection attacks: a data-driven approach to in-vehicle intrusion detection,'' in \emph{The Annual Conference on Cyber and Information Security Research}.\hskip 1em plus 0.5em minus 0.4em\relax ACM, 2017, p.~11.

\bibitem{he2016sdvn}
Z.~He, J.~Cao, and X.~Liu, ``\mbox{SDVN}: enabling rapid network innovation for heterogeneous vehicular communication,'' in \emph{IEEE network}, vol.~30, no.~4.\hskip 1em plus 0.5em minus 0.4em\relax IEEE, 2016, pp. 10--15.

\bibitem{singh2018ml}
P.~K. Singh, S.~K. Jha, S.~K. Nandi, and S.~Nandi, ``Ml-based approach to detect ddos attack in v2i communication under sdn architecture,'' in \emph{TENCON}.\hskip 1em plus 0.5em minus 0.4em\relax IEEE, 2018, pp. 0144--0149.

\bibitem{khan2019vehicle}
Z.~Khan, M.~Chowdhury, M.~Islam, C.-Y. Huang, and M.~Rahman, ``In-vehicle false information attack detection and mitigation framework using machine learning and software defined networking,'' \emph{arXiv:1906.10203}, 2019.

\bibitem{opendbc}
\mbox{CommaAI}, ``\mbox{OpenDBC},'' [Online]. 2019, \mbox{Available:} \url{https://github.com/commaai/opendbc}.

\bibitem{Goodfellow-et-al-2016}
I.~Goodfellow, Y.~Bengio, and A.~Courville, \emph{Deep Learning}.\hskip 1em plus 0.5em minus 0.4em\relax MIT Press, 2016, \url{http://www.deeplearningbook.org}.

\bibitem{kingma2014adam}
D.~P. Kingma and J.~Ba, ``Adam: A method for stochastic optimization,'' \emph{arXiv:1412.6980}, 2014.

\bibitem{GENI-Berman}
\mbox{B. Mark} \emph{et~al.}, ``\mbox{GENI}: A federated testbed for innovative network experiments,'' in \emph{Computer Networks}, 2014.

\bibitem{floodlight}
Floodlight, ``\mbox{Project Floodlight},'' [Online]. 2012, \mbox{Available:} \url{http://www.projectfloodlight.org/}.

\bibitem{art2018}
M.-I. Nicolae, M.~Sinn, M.~N. Tran, B.~Buesser, A.~Rawat, M.~Wistuba, V.~Zantedeschi, N.~Baracaldo, B.~Chen, H.~Ludwig, I.~Molloy, and B.~Edwards, ``Adversarial robustness toolbox v1.2.0,'' \emph{arXiv:1807.01069}, 2018.

\end{thebibliography}

\end{document}